\documentclass[10pt,journal,letterpaper,compsoc,twocolumn]{IEEEtran}

\usepackage{cite}
\usepackage{graphicx}
\usepackage{psfrag}
\usepackage{subfigure}
\usepackage{url}
\usepackage{stfloats}
\usepackage{amssymb,amsmath}
\usepackage{amsfonts}
\usepackage{array}

\begin{document}

\title{Higher-Order Markov Tag-Topic Models for Tagged Documents and Images}

\author{Jia Zeng, Wei Feng, William K. Cheung and Chun-Hung Li
\IEEEcompsocitemizethanks{\IEEEcompsocthanksitem
J. Zeng is with the School of Computer Science and Technology,
Soochow University, Suzhou 215006, China.
He is also with the Shanghai Key Laboratory of Intelligent Information Processing, China, and
the Center for Bioinformatics and Computational Biology, University of Maryland, College Park, MD 20742, USA.
To whom correspondence should be addressed.
E-mail: j.zeng@ieee.org.
\IEEEcompsocthanksitem
W. Feng is with the School of Computer Science and Technology, Tianjin University, China.
\IEEEcompsocthanksitem
W. K. Cheung and C.-H. Li are with the Department of Computer Science,
Hong Kong Baptist University, Kowloon Tong, Hong Kong.
}
}

\IEEEcompsoctitleabstractindextext{
\begin{abstract}
This paper studies the topic modeling problem of tagged documents and images.
Higher-order relations among tagged documents and images are major and ubiquitous characteristics,
and play positive roles in extracting reliable and interpretable topics.
In this paper,
we propose the tag-topic models (TTM) to depict such higher-order topic structural dependencies within the Markov random field (MRF) framework.
First,
we use the novel factor graph representation of latent Dirichlet allocation (LDA)-based topic models from the MRF perspective,
and present an efficient loopy belief propagation (BP) algorithm for approximate inference and parameter estimation.
Second,
we propose the factor hypergraph representation of TTM,
and focus on both pairwise and higher-order relation modeling among tagged documents and images.
Efficient loopy BP algorithm is developed to learn TTM,
which encourages the topic labeling smoothness among tagged documents and images.
Extensive experimental results confirm the incorporation of higher-order relations to be effective in enhancing the overall topic modeling performance,
when compared with current state-of-the-art topic models,
in many text and image mining tasks of broad interests such as word and link prediction,
document classification,
and tag recommendation.
\end{abstract}

}

\maketitle

\IEEEdisplaynotcompsoctitleabstractindextext

%
\IEEEpeerreviewmaketitle

\section{Introduction} \label{s1}

\begin{figure*}[t]
\centering
\includegraphics[width=1\linewidth]{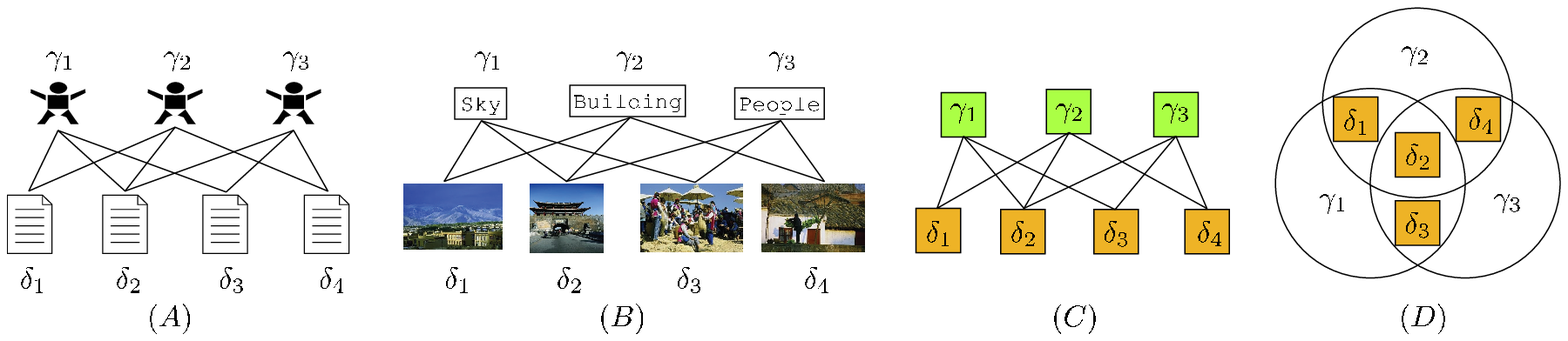}
\caption{Examples of tagged documents and images:
(A) the tagged documents with the tags being authors,
(B) the tagged images with the tags being annotations,
(C) the bipartite graph representation,
and
(D) the higher-order relation distinguishes the specific topic of $\delta_2$ from the combined topics of $\delta_1$, $\delta_3$, and $\delta_4$.
}
\label{tag}
\end{figure*}

The goal of this work is to model and infer semantically meaningful word clusters, referred to as topics,
from large-scale tagged documents and images.
In a broad sense,
we define a tag as a label that characterizes certain properties of documents and images.
For example,
the ``author" tag identifies the authorship of document,
and the ``time stamp" tag marks when the document is published.
On the other hand,
we can treat images as documents composed of visual words.
The users often manually annotate images by semantic tags such as ``building" or ``tree" to label local contents or objects of interests.
Generally,
one document may be associated with multiple tags,
and one tag may be attached to multiple documents.
Fig.~\ref{tag}A illustrates an example of tagged documents with the tags being authors,
where the link denotes that the author writes the document.
Fig.~\ref{tag}B shows another example of tagged images,
where four images are annotated with three tags ``sky", ``building", and ``people".
We can conveniently represent tagged documents and images by a bipartite graph in Fig.~\ref{tag}C,
which is composed of the tag nodes $\{\gamma\}$ and the document or image nodes $\{\delta\}$ connected by links.
In the bipartite graph,
tags often connect multiple documents or images so that build higher-order relations among documents or images.

Besides the pairwise relations,
the higher-order relations among tagged documents and images formed by multiple tags are major and ubiquitous characteristics.
For example,
the authors $\{\gamma_1,\gamma_2\}$ collaborate to write the document $\delta_1$ on the topic ``Machine Learning"
denoted by the intersection subset of two circles $\{\gamma_1,\gamma_2\}$ in Fig.~\ref{tag}D.
Similarly,
$\{\gamma_1,\gamma_3\}$ collaborate to write $\delta_3$ on the topic ``Computer Vision",
and $\{\gamma_2,\gamma_3\}$ collaborate to write $\delta_4$ on the topic ``Data Mining".
These three authors $\{\gamma_1,\gamma_2,\gamma_3\}$ also jointly collaborate to write the document $\delta_2$
denoted by the intersection subset of three circles $\{\gamma_1, \gamma_2, \gamma_3\}$ in Fig.~\ref{tag}D.
If we simply decompose the higher-order relation $\{\gamma_1, \gamma_2, \gamma_3\}$ into three pairwise relations
$\{\{\gamma_1,\gamma_2\},\{\gamma_1,\gamma_3\},\{\gamma_2,\gamma_3\}\}$,
we may just come to the conclusion that $\delta_2$ focuses on
the combined topics of ``Machine Learning", ``Computer Vision", and ``Data Mining".
Nevertheless,
the possibility that $\delta_2$ is in fact modeling a totally new topic like ``Computational Biology"
in the intersection subset of three circles will be excluded as shown in Fig.~\ref{tag}D.
Obviously,
$\delta_2$ lies in the specific subset of $\{\gamma_1, \gamma_2, \gamma_3\}$,
which is quite different from the union of subsets $\{\{\gamma_1,\gamma_2\},\{\gamma_1,\gamma_3\},\{\gamma_2,\gamma_3\}\}$.
So,
the explicit modeling the higher-order relation among documents constituted by multiple tags $\{\gamma_1, \gamma_2, \gamma_3\}$ is needed to
distinguish the specific topic of $\delta_2$ from the combined topics of $\delta_1$, $\delta_3$, and $\delta_4$.
Furthermore,
modeling such higher-order relations also reflects the truth that the tags $\{\gamma_1, \gamma_2, \gamma_3\}$ are often attached
to the document $\delta_2$ jointly rather than separately to explain the document content.
Similar higher-order relations among images induced by multiple tags can be also found in Fig.~\ref{tag}B.

However,
prior efforts at pairwise relation modeling in topic models~\cite{Blei:03,Erosheva:04,Nallapati:08,Chang:10,Gruber:08,Liu:09,Zeng:09,Hal:09,Dietz:07,Mei:08,Wang:09}
rarely consider higher-order relations that may encode specific topic structural dependencies among tagged documents and images.
Therefore,
in this paper,
we propose the tag-topic models (TTM) to describe such higher-order topic structural dependencies within the Markov random field (MRF) framework.
This approach extends our previous work in modeling higher-order relations of coauthors~\cite{Zeng:10} to the more generic tagged documents and images,
allowing us to develop more efficient inference and parameter estimation algorithms within the theoretically well-founded MRF framework.

First,
we reformulate the topic modeling task as a labeling problem from the novel MRF perspective.
We represent the latent Dirichlet allocation~\cite{Blei:03} (LDA)-based topic models as factor graphs~\cite{Kschischang:01,Bishop:book},
and develop the classic loopy belief propagation (BP) algorithm to make approximate inference and parameter estimation.
Second,
we represent TTM using the factor hypergraph~\cite{Klamt:09} according to the bipartite graph in Fig.~\ref{tag}C,
and focus on both pairwise and higher-order relation modeling within the higher-order MRF framework.
Indeed,
such higher-order MRF has recently found important applications in modeling high-level image structural priors in many computer vision problems,
including image restoration, disparity estimation and object segmentation~\cite{Lan:06,Feng:09}.

Generally,
inferring the higher-order MRF is intrinsically a computationally expensive problem since
even encoding $M$-order topic structural dependencies of $J$ topics requires $J^M$ labeling configurations.
However,
similar to image structural priors,
higher-order relations used in topic modeling also have certain properties such as {\em smoothness} or {\em sparsity}~\cite{Li:book,Szeliski:08,Feng:09},
which makes them easy to handle.
Intuitively,
the co-tagged documents and images tend to have a higher likelihood to share the similar topic labeling configuration.
Based on the smoothness or sparsity prior,
many higher-order topic labeling configurations are equally unlikely and thus need not to be encouraged.
Therefore,
we encourage only a total of $JM$ smooth topic labeling configurations in TTM,
which avoids encoding $J^M$ arbitrary topic structural dependencies.
To this end,
we design the higher-order functions to encode the major and representative smoothness relations,
and develop the loopy BP algorithm~\cite{Bishop:book,Lan:06} to make efficient inference and parameter estimation of TTM.

The rest of this paper is organized as follows.
Section~\ref{s2} introduces related work.
Sections~\ref{s3} presents MRF for topic modeling and develops loopy BP algorithms for approximate inference and parameter estimation.
Section~\ref{s4} proposes TTM and focuses on pairwise and higher-order relation modeling among tagged documents and images.
Section~\ref{s5} shows extensive experimental results on several challenging text and image mining tasks of broad interests.
Finally,
Section~\ref{s6} draws conclusions and envisions future work.

\section{Related Work} \label{s2}

\begin{figure*}[t]
\centering
\includegraphics[width=1\linewidth]{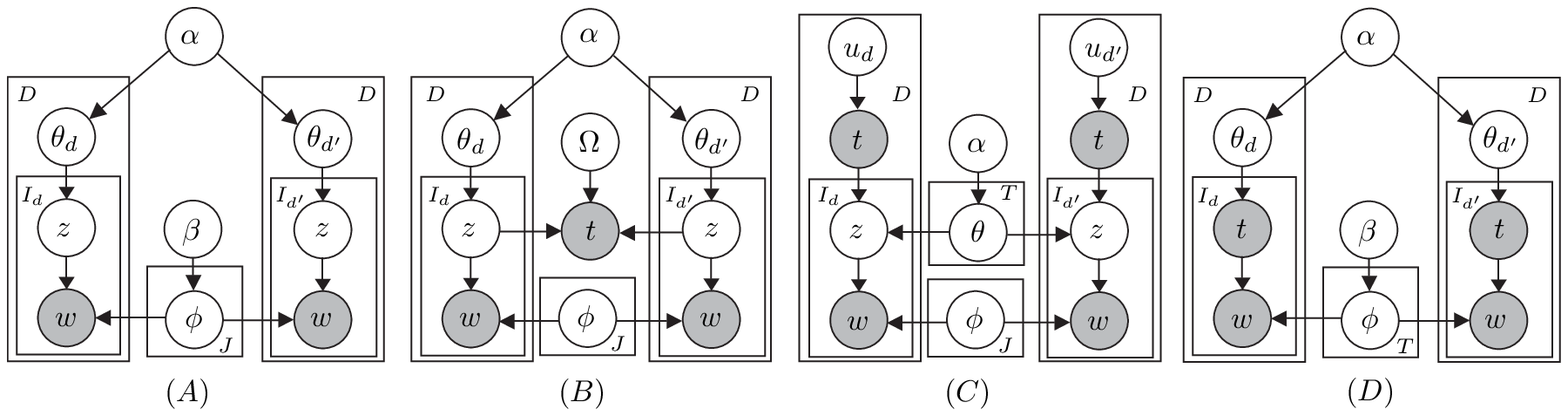}
\caption{
(A) LDA~\cite{Blei:03},
(B) RTM~\cite{Chang:10},
(C) ATM~\cite{Rosen-Zvi:04},
and (D) L-LDA~\cite{Ramage:09}.
The shaded circles are observed variables,
while others are latent variables and parameters.
The plates indicate replication.
The hyperparameter $\beta$ in (B)-(C) and some subscripts are omitted for simplicity.
}
\label{ptm}
\end{figure*}

The use of probabilistic topic models for text mining is the state-of-the-art approach such as learning terminological ontologies~\cite{Wei:10}.
LDA~\cite{Blei:03} is the basic topic model in Fig.~2A.
It allocates a topic label $z$ to each word $w$ in the document $d$
based on the document-specific topic proportion $\theta_d$ and the topic-specific multinomial distribution $\phi$ over vocabulary words,
where $\theta_d$ and $\phi$ are smoothed by two conjugate Dirichlet hyperparameters $\alpha$ and $\beta$,
respectively.
The plates indicate replication.
For example,
the document repeats $D$ times in the corpus,
the word repeats $I_d$ times in the document $d$,
and there are a total of $J$ topics.
LDA builds implicit links between two documents $\{d,d'\}$ by sharing the same topic distribution $\phi$,
and it encourages similar topic labeling configurations if two documents contain similar words.
However,
LDA uses only $\phi$ to exchange topic information among documents,
but ignores the rich link information like citations or hyperlinks between documents.
This motivates the recent variants of LDA that regularize topic distribution $\phi$ through pairwise relations between documents.

Pairwise topic models focus on the link generation process,
which in turn influences the topic allocation for words.
Link LDA~\cite{Erosheva:04} uses the document-specific topic proportion and
the topic-specific distribution over documents to generate a cited document by the citing document.
When two documents cite the same document,
they tend to have similar topic labeling configuration over words.
In this sense,
link LDA indirectly depicts a co-citation link between documents citing the same other documents,
and scales badly to the large-scale corpus because its parameters increases with the total number of documents.
To overcome these weaknesses,
pairwise LDA~\cite{Nallapati:08} directly generates the binary citation link variable between two documents using a topic-dependent Bernoulli distribution.
But it randomly uses one of topic labels rather than the entire topic labels in the document to generate links,
which significantly limits the influence of link information on topic regularization.
To relax this limitation,
the relational topic model (RTM) (Fig.~\ref{ptm}B)~\cite{Chang:10} represents entire document topics by the mean value of the document topic proportions.
It then uses the Hadamard product of mean values $\overline{z}_d \circ \overline{z}_{d'}$ from two linking documents $\{d,d'\}$ as link features,
which are learned by the generalized linear model $\Omega$ to generate the observed citation link variable $t=1$.
If the citation link variables are replaced by the observed tags,
RTM can be adapted to account for the tagged documents and images.
Similar to the basic idea of RTM,
latent topic hypertext model~\cite{Gruber:08} assumes that links originate from words and
uses partial word topic labels to generate links.

Furthermore,
topic-link LDA~\cite{Liu:09}, multirelational topic models~\cite{Zeng:09}, and Markov random topic fields~\cite{Hal:09}
simultaneously generate multiple types of links such as citations, coauthor relations, and social community of authors to improve the accuracy of topic modeling.
Citation influence model~\cite{Dietz:07} allows that the topic of a citing document is dependent on
either its own topic proportions or its cited documents' topic proportions.
Topic modeling with network regularization~\cite{Mei:08} adopts a graph-based regularizer to
encourage the minimum Euclidean distance between document-layer topic labeling configurations.
Markov topic models~\cite{Wang:09} use the Gaussian Markov random field
to describe topic interactions among documents in different conferences.
Nevertheless,
all recent pairwise topic models have limited expressive power,
because they are insufficient to depict higher-order relations among tagged documents and images.

Author-topic models (ATM)~\cite{Rosen-Zvi:04} (Fig.~\ref{ptm}C) and labeled LDA (L-LDA)~\cite{Ramage:09} (Fig.~\ref{ptm}D)
are able to associate observed tags with words directly.
ATM uses a document-specific uniform distribution $u_d$ to generate a tag,
and further uses the tag-specific topic proportions $\theta_t$ to generate a topic label for the word.
The plate on $\theta$ indicates that there are $T$ unique tags.
All documents with the tag $t$ will share the same tag-specific topic proportions $\theta_t$,
which implicitly encodes the pairwise relation of documents associated with the tag $t$.
L-LDA constrains latent topics to be observed tags generated from the document-specific topic proportions $\theta_d$ over tags,
where each tag is further associated with a multinomial distribution $\phi_t$ to generate words.
In this sense,
L-LDA is a supervised topic model because it replaces latent topic labels as observed tags.
All documents with the tag $t$ will share the same multinomial distribution $\phi_t$,
which also encodes statistical information of documents associated with the tag $t$.
However,
the higher-order relations among documents and images due to multiple connected tags have been largely neglected in both ATM and L-LDA,
which motivates us to explore a more specific higher-order TTM in this study.

\section{MRF for Topic Modeling} \label{s3}

\subsection{The Labeling Problem}

\begin{table}[t]
\centering
\caption{Notations}
\begin{tabular}{|l|l|} \hline
$1 \le d \le D$                      &Document index              \\ \hline
$1 \le w \le W$                      &Vocabulary word index       \\ \hline
$1 \le t \le T$                      &Tag index                   \\ \hline
$1 \le j \le J$                      &Topic index                 \\ \hline
$\mathbf{w} = \{w,d\}$               &Bag of words                \\ \hline
$\mathbf{z} = \{z_{w,d}\}$           &Topic labels for words      \\ \hline
$\mathbf{z}_{-w,d}$                  &Labels for $d$ excluding $w$   \\ \hline
$\mathbf{z}_{w,-d}$                  &Labels for $w$ excluding $d$   \\ \hline
$\theta_d$                           &Factor of document $d$ \\ \hline
$\phi_w$                             &Factor of word $w$     \\ \hline
$\gamma_t$                           &Factor of tag $t$      \\ \hline
$\delta_d$                           &Factor hyperedge        \\ \hline
$\mu(z_{w,d})$                       &Topic messages              \\ \hline
$\mu(\mathbf{z}_{\cdot,d})$ and $\mu(\mathbf{z}_{w,\cdot})$    &$\sum_w \mu(z_{w,d})$ and $\sum_d \mu(z_{w,d})$       \\ \hline
$f(\cdot)$                           &Factor functions            \\ \hline
$\alpha,\beta$                       &Dirichlet hyperparameters   \\ \hline
\end{tabular}
\label{notation}
\end{table}

Table~\ref{notation} summarizes some important notations in this paper.
From a new perspective,
this subsection formulates the topic modeling as the labeling problem within the MRF framework.
The objective of topic modeling is to assign a set of semantic topic labels $\mathbf{z} = \{z_{w,d}\}$ to
explain the observed bag of words $\mathbf{w} = \{w,d\}$,
where $1 \le w \le W$ is the word index in the vocabulary and $1 \le d \le D$ is the document index in the corpus.
Generally,
the topic label $z_{w,d}$ takes one of the topic index $1 \le j \le J$ and partitions all words into $J$ topic groups,
so that the topic modeling technique is often viewed as one of the word clustering paradigms.
In theory,
MRF solves the labeling problem by assigning the best topic labels according to the {\em maximum a posteriori} (MAP) estimation,
and this MRF-MAP framework has found many important applications in image analysis and computer vision~\cite{Li:book}.
More specifically,
MRF attempts to find the best topic labeling configuration $\mathbf{z}$ over words $\mathbf{w}$ through maximizing the posterior probability $p(\mathbf{z}|\mathbf{w})$,
which is in nature a prohibited combinatorial optimization problem in the discrete latent topic space.
To avoid the high computational cost,
MRF often uses {\em smoothness} or {\em sparsity} property of the labeling problem to reduce the total number of possible labeling configurations~\cite{Li:book,Szeliski:08,Feng:09}.

As far as topic modeling is concerned,
we only encourage smoothness of neighboring topic labels,
i.e.,
the neighboring topic labels tend to be the same.
To this end,
we define the neighborhood system of the topic label $z_{w,d}$ as $\mathbf{z}_{-w,d}$ and $\mathbf{z}_{w,-d}$,
where $\mathbf{z}_{-w,d}$ denotes a set of topic labels associated with all word indices in the document $d$ excluding the word index $w$,
and $\mathbf{z}_{w,-d}$ denotes a set of topic labels associated with the word index $w$ in all documents excluding $d$.
Furthermore,
we use the factor graph~\cite[Chapter~8.4.3]{Kschischang:01,Bishop:book} to represent LDA,
and treat parameters $\theta$ and $\phi$ as factors with parameterized functions~\cite{Bishop:book}.
By designing the proper factor functions,
which are equivalent to clique potentials,
we can encourage or penalize different local labeling configurations in the neighborhood system.
More specifically,
we encourage the topic labeling smoothness among $\{z_{w,d}, \mathbf{z}_{-w,d}, \mathbf{z}_{w,-d}\}$.

\begin{figure*}[t]
\centering
\includegraphics[width=1\linewidth]{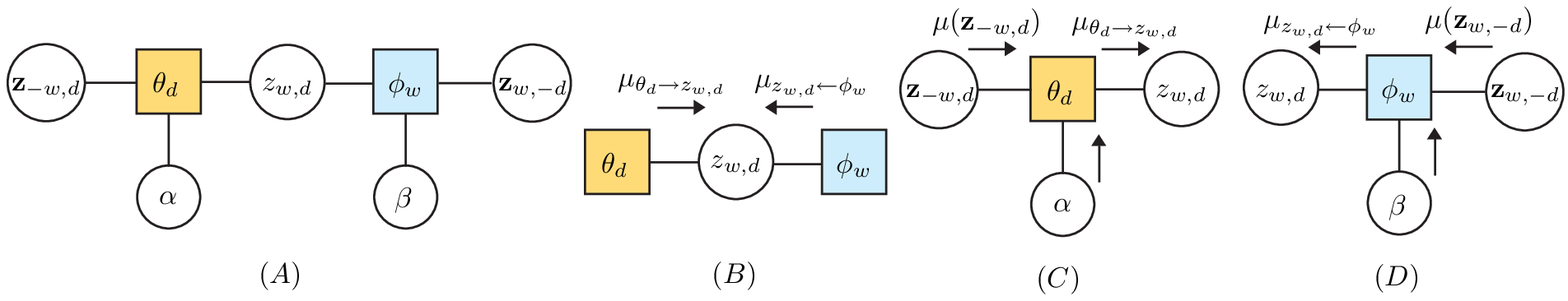}
\caption{(A) Factor graph for LDA.
(B) Passing messages from the factors $\theta_d$ and $\phi_w$ to the variable $z_{w,d}$.
(C) and (D) Passing messages from the neighboring variables $\mathbf{z}_{-w,d}$ and $\mathbf{z}_{w,-d}$ to factors $\theta_d$ and $\phi_w$, respectively.
Arrows show the directions of message passing.
}
\label{factor}
\end{figure*}

In this paper,
we consider a type of LDA with fixed symmetric Dirichlet hyperparameters $\alpha$ and $\beta$~\cite{Griffiths:04}
in order to avoid the complex full Bayesian inference of $\alpha$ and $\beta$, respectively.
We transform the generative graphical representation of LDA in Fig.~\ref{ptm}A to
the factor graph~\cite{Kschischang:01,Bishop:book} in Fig.~\ref{factor}A from the MRF perspective.
We illustrate the factors $\theta_d$ and $\phi_w$ as squares,
and denote their connected variables $z_{w,d}$ as circles.
Obviously,
the factors $\theta_d$ and $\phi_w$ connects the neighboring topic labeling configurations $\{z_{w,d}, \mathbf{z}_{-w,d}, \mathbf{z}_{w,-d}\}$.
In this way,
the hierarchically directed graphical model of LDA in Fig.~\ref{ptm}A becomes a more generic undirected graphical model in Fig.~\ref{factor}A.
We absorb the observed word index $w$ as the index of the factor $\phi_w$,
which is similar to absorbing the observed document index $d$ as the index of the factor $\theta_d$ in Fig.~\ref{ptm}A.
Because the factors can be parameterized functions~\cite{Bishop:book},
both $\theta_d$ and $\phi_w$ can be the same multinomial functions smoothed by the Dirichlet priors defined in LDA~\cite{Griffiths:04}.
Also,
both hyperparameters can be viewed as pseudo-counts in estimating the corresponding multinomial distributions.
This resembles the collapsed GS~\cite{Griffiths:04} that integrates out parameter variables $\theta$ and $\phi$
and treats hyperparameters $\alpha$ and $\beta$ as pseudo topic counts in order to perform inference on the collapsed hidden variable space $\mathbf{z}$.

Recently,
LDA has been reformulated as a Bayesian network~\cite{Heinrich:08},
which is one of the constrained undirected graphical models (MRF) with causal dependencies between hidden variables.
Indeed,
Fig.~\ref{ptm}A and Fig.~\ref{factor}A reflect two facets of LDA,
in which the former focuses more on the generative process of the observed words hierarchically,
while the latter emphasizes more on the topic labeling smoothness within the MRF framework.

The original factor graph representation for MRF~\cite{Kschischang:01} can be naturally extended to describe the generative process of a probabilistic model.
For example,
one extension is the directed factor graph~\cite{Dietz:10} that enhances the visual language to represent LDA.
Because the topic modeling task can be formulated as the labeling problem from the MRF perspective,
the original undirected factor graph has enough expressive power to represent LDA directly.
Although the undirected graph does not explicitly emphasize the generative process as the directed counterpart~\cite{Dietz:10},
it still captures the underlying structural dependencies of hidden variables without loss of information.
In this sense,
the factor graph may be a more generic visual representation for both directed and undirected graphical models in various real-world applications.

Although the factor graph in Fig.~\ref{factor}A is slightly different from the directed graphical model in Fig.~\ref{ptm}A,
it can fulfill the same topic modeling task using the specific neighborhood systems and factor functions.
First,
both Figs.~\ref{factor}A and~\ref{ptm}A have the same neighborhood system because the connection of hidden variables remains the same.
Second,
in the next subsection,
we shall design specific factor functions to realize the same topic modeling goal as Fig.~\ref{factor}A without loss of information.

\subsection{Inference and Parameter Estimation} \label{s3.2}

The loopy BP~\cite[Chapter 8]{Bishop:book} algorithms such as the sum-product and max-sum algorithms provide efficient
and approximate solutions to inference problems for graphs with loops in Fig.~\ref{factor}A.
Rather than directly calculating the posterior probability $p(\mathbf{z}|\mathbf{w})$,
we turn to calculating the posterior marginal probability $p(z_{w,d})$,
referred to as the {\em message} $\mu(z_{w,d})$,
which can be normalized efficiently using a local computation.
Message passing proceeds from variables to factors,
and in turn from factors to variables until convergence after several iterations.
In this subsection,
we adopt the sum-product algorithm to infer the marginal posterior probability $\mu(z_{w,d})$.

The message passing scheme is an instantiation of the E-step of expectation-maximization (EM) algorithm~\cite{Bishop:book},
which has been widely used to infer the marginal probabilities of hidden variables
in various graphical models according to the maximum-likelihood estimation.
For example,
the E-step inference for Gaussian mixture models (GMM)~\cite{Zeng:08c},
the forward-backward algorithm for hidden Markov models (HMM)~\cite{Zeng:06a},
and the probabilistic relaxation labeling algorithm for MRF~\cite{Zeng:08a}
can be all formulated within the message passing framework~\cite{Frey:05,Werner:07,Bishop:book}.
After the E-step,
we can estimate parameters $\theta$ and $\phi$ based on the inferred marginal probabilities $\mu(z_{w,d})$ at the M-step of EM algorithm,
which is almost the same as those EM algorithms for learning other finite mixture models like GMM and GMM-based HMM.
More details on learning finite mixture models using the EM algorithm can be found in the book~\cite{Bishop:book}.

Fig.~\ref{factor}B shows the message passing from two factors to the variable.
The message $\mu(z_{w,d})$ is the product of both input messages,
\begin{align} \label{f2v}
\mu(z_{w,d}) \propto \mu_{\theta_d \rightarrow z_{w,d}}(z_{w,d}) \times \mu_{\phi_w \rightarrow z_{w,d}}(z_{w,d}),
\end{align}
where we use the arrows to denote the message passing directions.
The normalized message $\mu(z_{w,d})$ is in turn passed back to the factors.
In Fig.~\ref{factor}C and~\ref{factor}D,
the messages from factors to variables can be further calculated based on all input messages from neighboring variables as follows,
\begin{align} \label{v2f1}
\mu_{\theta_d \rightarrow z_{w,d}}(z_{w,d}) = \sum_{\mathbf{z}_{-w,d}} f_{\theta_d} \prod_{-w} \mu(z_{-w,d}) \alpha, \\
\label{v2f2}
\mu_{\phi_w \rightarrow z_{w,d}}(z_{w,d}) = \sum_{\mathbf{z}_{w,-d}} f_{\phi_w} \prod_{-d} \mu(z_{w,-d}) \beta,
\end{align}
where $\mathbf{z}_{-w,d}$ and $\mathbf{z}_{w,-d}$ represent all possible neighboring labeling configurations of $z_{w,d}$,
and $f(\cdot)$ is the factor function that evaluates the topic structural dependencies of input topic messages.
The topic labeling smoothness prior implies that only $J$ topic configurations are encouraged in~\eqref{v2f1} and~\eqref{v2f2}.
Thus,
we can rewrite~\eqref{v2f1} and~\eqref{v2f2} as
\begin{align} \label{v2f3}
\mu_{\theta_d \rightarrow z_{w,d}}(z_{w,d}=j) = f_{\theta_d} \prod_{-w} \mu(z_{-w,d}=j) \alpha, \\
\label{v2f4}
\mu_{\phi_w \rightarrow z_{w,d}}(z_{w,d}=j) = f_{\phi_w} \prod_{-d} \mu(z_{w,-d}=j) \beta.
\end{align}
In practice,
Eqs.~\eqref{v2f1} and~\eqref{v2f2} often cause the product of multiple input messages close to zero~\cite{Zeng:08}.
To avoid arithmetic underflow,
we approximate the product of messages by the sum of messages because the product value increases when the sum value increases,
\begin{align}
\label{p2s1}
\prod_{-w} \mu(z_{-w,d}) \alpha \propto \sum_{-w} \mu(z_{-w,d}) + \alpha, \\
\label{p2s2}
\prod_{-d} \mu(z_{w,-d}) \beta \propto \sum_{-d} \mu(z_{w,-d}) + \beta.
\end{align}
Such approximations as~\eqref{p2s1} and~\eqref{p2s2} transform the standard sum-product to the sum-sum algorithm,
which is still good at passing messages in MRF with acceptable performance~\cite{Werner:07,Zeng:08}.

For convenience we use the shorthand notations $\mu(\mathbf{z}_{\cdot,d}) = \sum_w \mu(z_{w,d})$,
$\mu(\mathbf{z}_{w,\cdot}) = \sum_d \mu(z_{w,d})$,
$\mu(\mathbf{z}_{-w,d}) = \sum_{-w} \mu(z_{-w,d})$,
and $\mu(\mathbf{z}_{w,-d}) = \sum_{-d} \mu(z_{w,-d})$
in the subsequent formulas.
In MRF~\cite{Bishop:book},
the factor functions $f(\cdot)$ correspond to the clique potentials,
which can be designed arbitrarily to encode our prior knowledge on encouraging or penalizing topic labeling configurations.
Indeed,
the higher value of $f(\cdot)$ encourages passing more neighboring messages.
Here,
we design $f_{\theta_d}$ and  $f_{\phi_w}$ as
\begin{align} \label{ftheta}
f_{\theta_d} = \frac{1}{\sum_j [\mu(\mathbf{z}_{-w,d}=j) + \alpha]}, \\
\label{fphi}
f_{\phi_w} = \frac{1}{\sum_w [\mu(\mathbf{z}_{w,-d}=j) + \beta]}.
\end{align}
Eq.~\eqref{ftheta} normalizes the input messages by the total number of topics associated with the document $d$
in order to make output messages comparable across different documents.
Eq.~\eqref{fphi} normalizes the input messages by the total number of messages of all word indices in the vocabulary
in order to make output messages comparable across different vocabulary words.

Combining~\eqref{f2v} to \eqref{fphi} yields the complete message update equation,
\begin{align} \label{message}
\mu(z_{w,d}=j) \propto \frac{\mu(\mathbf{z}_{-w,d}=j) + \alpha}
{\sum_j [\mu(\mathbf{z}_{-w,d}=j) + \alpha]} \notag \\
\times \frac{\mu(\mathbf{z}_{w,-d}=j) + \beta}{\sum_w [\mu(\mathbf{z}_{w,-d}=j) + \beta]},
\end{align}
where the notations $-w$ and $-d$ denote all word indices except $w$ and all document indices except $d$,
and the notations $\mu(\mathbf{z}_{-w,d})$ and $\mu(\mathbf{z}_{w,-d})$ represent the sum of all possible neighboring messages excluding the current message $\mu(z_{w,d})$.
We normalize the updated message locally in terms of $j$ so that $\sum_j \mu(z_{w,d} = j) = 1$.

In practice,
after finite $N$ iterations,
the message will converge in the factor graph as shown in Fig.~\ref{factor}A.
BP usually converges fast with $N \le 500$.
Note that we need to multiply the number of word counts or the relative word frequencies to the corresponding word topic message during message passing and parameter estimation.

Given the inferred marginal posterior probability $\mu(z_{w,d})$,
the parameter estimation of $\theta$ and $\phi$ can be performed simply using~\eqref{v2f3} and~\eqref{v2f4} (Figs.~\ref{factor}C and~\ref{factor}D)
by adding all input messages including $\mu(z_{w,d})$ evaluated by the corresponding factor functions,
\begin{align}
\label{thetad}
\theta_d(j) = \frac{\mu(\mathbf{z}_{\cdot,d}=j) + \alpha}{\sum_j [\mu(\mathbf{z}_{\cdot,d}=j) + \alpha]}, \\
\phi_w(j) = \frac{\mu(\mathbf{z}_{w,\cdot}=j) + \beta}{\sum_w [\mu(\mathbf{z}_{w,\cdot}=j) + \beta]}.
\label{phiw}
\end{align}

Alternatively,
we can also derive the parameter estimation equations using the EM algorithm~\cite{Bishop:book}.
In the E-step,
we calculate the marginal posterior probability $\mu(z_{w,d})$.
Employing the multinomial-Dirichlet conjugacy and Bayes' rule,
we get the following marginal Dirichlet distributions~\cite{Heinrich:08},
\begin{align}
\label{theta}
p(\theta_d) = \text{Dir}(\theta_d|\mu(\mathbf{z}_{\cdot,d})+\alpha), \\
p(\phi_w) = \text{Dir}(\phi_w|\mu(\mathbf{z}_{w,\cdot})+\beta),
\label{phi}
\end{align}
In the M-step,
maximizing~\eqref{theta} and~\eqref{phi} with respect to $\theta_d$ and $\phi_w$ also results in the same parameter estimation equations~\eqref{thetad} and~\eqref{phiw}.

\subsection{Discussion} \label{s3.3}

LDA is the hierarchical Baysian model that maximizes the objective $p(\mathbf{w},\mathbf{z})$ to generate topic labels for words,
while MRF is the undirected model that maximizes the objective $p(\mathbf{z}|\mathbf{w})$ to assign the best topic labels to words.
Their objectives are identical according to the Bayes' rule $p(\mathbf{z}|\mathbf{w}) \propto p(\mathbf{w},\mathbf{z})$ since $p(\mathbf{w})$ is a constant in terms of $\mathbf{z}$.
The collapsed Gibbs sampling (GS)~\cite{Griffiths:04} and variational Bayes (VB)~\cite{Blei:03} have been two commonly-used approximate inference algorithms for LDA-based topic models.
In this paper,
we provide an alternative inference method for LDA-based topic models using the BP algorithm from the novel MRF perspective.

GS resembles the proposed BP except that it randomly samples a topic label from the marginal posterior probability $\mu(z_{w,d})$ for each word token,
and immediately updates parameters based on the currently sampled topic label.
Therefore,
GS needs to sample a topic label for each word token in the document,
while BP only calculates the message for each word index in the vocabulary within the document.
Due to the word sparsity in the document,
BP significantly lowers the computational cost than GS.
In addition,
the randomly sampled topic label in GS always loses some information compared with the marginal posterior probability in BP.
As a result,
BP is more accurate than GS in parameter estimation because it keeps and uses the complete messages at each learning iteration without loss of information.

VB uses the Jensen's inequality to get an adjustable lower bound on the objective function,
and maximizes the objective through maximizing the lower bound by tuning a set of variational parameters.
VB also resembles the proposed BP except that it calculates the topic messages by minimizing the Kullback-Leibler (KL)
divergence between the variational distribution and the true posterior distribution.
Thus,
the variational message update equations in VB differ significantly from those in BP by involving the more complicated digamma functions.

For each learning iteration,
both BP and VB have the same computational cost $\mathcal{O}(JDW_d)$,
but GS requires $\mathcal{O}(JDI_d)$,
where $W_d$ is the average vocabulary size and $I_d$ is the average number of word tokens per document.
Because in a document the number of word indices is usually much smaller than the total number of word tokens due to the word sparsity,
i.e.,
$W_d \ll I_d$,
BP and VB generally scale much better than GS to the large-scale corpus.
More detailed comparisons among VB, GS and BP can be found in~\cite{Zeng:11}.

\section{Tag-Topic Models} \label{s4}

\subsection{Factor Hypergraph Representation}

\begin{figure*}[t]
\centering
\includegraphics[width=1\linewidth]{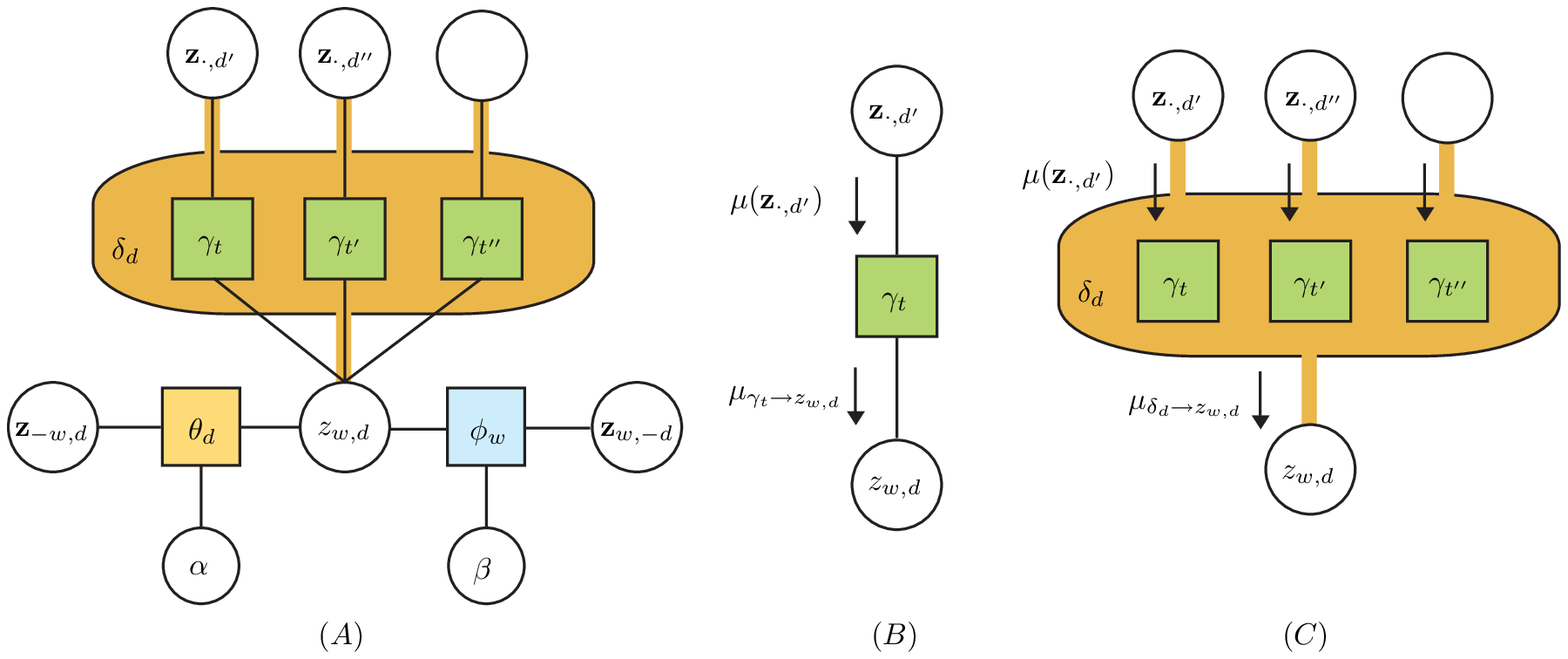}
\caption{(A) Factor hypergraph for tag-topic models,
where three tags are illustrated for simplicity.
(B) Pairwise relation modeling: passing messages through the factor $\gamma_t$.
(C) Higher-order relation modeling: passing messages through the factor hyperedge $\delta_d$ denoted by the yellow block.
Arrows show the directions of message passing.
}
\label{ttm}
\end{figure*}

Fig.~\ref{ttm}A shows the factor hypergraph representation of TTM,
which directly combines the factor graph in Fig.~\ref{factor}A with the bipartite graph in Fig.~\ref{tag}C.
Note that the undirected hypergraph is equivalent to the bipartite graph in Fig.~\ref{tag}C~\cite{Klamt:09},
where the factor hyperedge $\delta_d$ (denoted by the yellow block) connects the tag factors $\gamma_t$ attached to the document $d$ in Fig.~\ref{ttm}C.
In this factor hypergraph representation,
we absorb the observed tag $t$ as the index of the factor $\gamma_t$,
which connects the variable $z_{w,d}$ with its neighbors $\mathbf{z}_{\cdot,d'}$ using the solid black line as shown in Fig.~\ref{ttm}B.
We assume that the document pair $\{d,d'\}$ share the same tag $t$,
the document pair $\{d,d''\}$ share the same tag $t'$,
the the document $d$ are associated with three tags $\{t,t',t''\}$.
Although Fig.~\ref{ttm}A does not follow the standard definition of factor graphs due to the factor hyperedge $\delta_d$,
this variant of factor graph can represent both pairwise and higher-order relations among tagged documents and images as shown in Fig.~\ref{ttm}B and Fig.~\ref{ttm}C.
For example,
the topic labeling configurations $\mathbf{z}_{\cdot,d'}$ or $\mathbf{z}_{\cdot,d''}$ can influence its neighboring label $z_{w,d}$ separately
through the factor $\gamma_t$ or $\gamma_{t'}$ based on the pairwise relation in Fig.~\ref{ttm}B.
In the meanwhile,
$\{\mathbf{z}_{\cdot,d'},\mathbf{z}_{\cdot,d''}\}$ can also influence their neighbor $z_{w,d}$ jointly through the factor hyperedge $\delta_d$
based on the higher-order relation resulted from the connected tag factors $\{\gamma_t, \gamma_{t'}\}$ in Fig.~\ref{ttm}C.
As a result,
the factor function $f_{\gamma_t}$ encodes the pairwise relation between $\{\mathbf{z}_{\cdot,d'}, z_{w,d}\}$,
while the factor function $f_{\delta}$ depicts the higher-order relation among $\{\mathbf{z}_{\cdot,d'}, \mathbf{z}_{\cdot,d''}, z_{w,d}\}$.

\subsection{Credit Attribution}

Each attached semantic tag usually accounts for parts of words in documents or local contents in images.
The {\em credit attribution} task is to associate individual words in a document with their most appropriate tags~\cite{Ramage:09}.
In the probabilistic framework,
we assume that all tags in the document associate the same word with different likelihoods,
which can be calculated based on the pairwise relation formed by each tag $\gamma_t$ as shown in Fig.~\ref{ttm}B.
Specifically,
if $x_{w,d}$ is the tag label associated with the word $w$ in the document $d$,
we calculate the likelihood $p(x_{w,d} = t)$ based on the following similarity in terms of topic messages,
\begin{align} \label{credit}
p(x_{w,d}=t) = \sum_{j=1}^J \mu(z_{w,d} = j)\mu_{\gamma_t \rightarrow z_{w,d}}(z_{w,d} = j),
\end{align}
where the message $\mu_{\gamma_t \rightarrow z_{w,d}}(z_{w,d} = j)$ from the factor $\gamma_t$ will be introduced in the next subsections.
Intuitively,
Eq.~\eqref{credit} measures the similarity between the tag and the word content in the latent topic space.
In practice,
we randomly initialize $p(x_{w,d} = t)$ for all tags per word,
and iteratively update and normalize it using~\eqref{credit}.

\subsection{Pairwise and Higher-order Relation Modeling}

In Fig.~\ref{ttm}B,
the factor function $f_{\gamma_t}$ describes the pairwise topic dependencies between all pairs of documents connected with the tag $t$,
\begin{align} \label{pairwise}
f_{\gamma_t} = \frac{\sum_{d,d' \in \text{ne}(t)} \overline{\mu}_{d,t} \circ \overline{\mu}_{d',t'}}{|d,d' \in \text{ne}(t)|},
\end{align}
where the $\circ$ notation denotes the Hadamard (element-wise) product of two vectors~\cite{Chang:10},
the $\text{ne}(t)$ notation denotes the set of all connected documents with the tag $t$,
the $|d,d' \in \text{ne}(t)|$ notation indicates the total number of document pairs connected with the tag $t$,
and
\begin{align} \label{dt}
\overline{\mu}_{d,t} = \frac{\sum_w\mu(z_{w,d})p(x_{w,d} = t)}{\sum_w p(x_{w,d} = t)},
\end{align}
where $p(x_{w,d} = t)$ is defined in~\eqref{credit}.
The Hadamard product captures the similarity between two connected documents with the tag $t$ in the latent topic representations.
As a result,
Eq.~\eqref{pairwise} is the average Hadamard product of all pairs of documents connected with the tag $\gamma_t$
that encodes the dominant pairwise topic structural dependencies.
Eq.~\eqref{dt} is the weighted sum of all word messages in the document $d$ with respect to the tag $t$,
which can be viewed as the normalized message passed from all words in the document $d$ to the tag $t$.

In Fig.~\ref{ttm}C,
the factor function $f_{\delta_d}$ depicts the higher-order topic dependencies among documents and images through connected tags $\{t, t', t''\} \in \text{ne}(d)$.
Generally,
modeling the 3-order and 4-order relations is sufficient in practice because most documents and images often contain less than four tags in Table~\ref{datasets}.
Without loss of generality,
here we present the 3-order relation modeling,
and the higher than 3-order relation can be modeled similarly.
We design $f_{\delta_d}$ for 3-order relation based on the Hadmard product as follows,
\begin{align} \label{3order}
f_{\delta_d} = \frac{\sum_{d, d', d'' \in \text{ne}(t,t',t'')} \overline{\mu}_{d,t} \circ \overline{\mu}_{d',t'} \circ \overline{\mu}_{d'',t''}}
{|d, d', d'' \in \text{ne}(t,t',t'')|},
\end{align}
where the $|d, d', d'' \in \text{ne}(t,t',t'')|$ notation denotes
the total number of 3-order document or image triples constituted by the connected tags $t$, $t'$ and $t''$, respectively.
Obviously,
Eq.~\eqref{3order} is the average Hadmard product of all triples of documents or images,
capturing the major and representative 3-order topic structural dependencies among tagged documents and images.

\subsection{Inference and Parameter Estimation}

TTM in Fig.~\ref{ttm}A contains loops so that we develop the loopy BP algorithm~\cite{Bishop:book,Lan:06} for approximate inference and parameter estimation.
In subsection~\ref{s3.2},
we have calculated the messages $\mu_{\theta_d \rightarrow z_{w,d}}$ and $\mu_{\phi_w \rightarrow z_{w,d}}$ in~\eqref{v2f3} and~\eqref{v2f4}.
In this subsection,
we focus on computing the message $\mu_{\gamma_t \rightarrow z_{w,d}}$ and $\mu_{\delta_d \rightarrow z_{w,d}}$ based on the sum-product algorithm,
which involves not only pairwise relations but also higher-order relations among documents in Figs.~\ref{ttm}B and~\ref{ttm}C, respectively.
Under the topic smoothness constraint,
the message from the factor $\gamma_t$ to the variable $z_{w,d}$ is
\begin{align} \label{gamma2z}
\mu_{\gamma_t \rightarrow z_{w,d}} = f_{\gamma_t} \sum_{m \in \text{ne}(t) \backslash d} \overline{\mu}_{m,t},
\end{align}
where $m \in \text{ne}(t) \backslash d$ denotes all connected document with the tag $t$ except the current document $d$ in Fig.~\ref{ttm}B.
Similarly,
the message from the factor hyperedge $\delta_d$ to the variable $z_{w,d}$ is
\begin{align} \label{delta2z}
\mu_{\delta_d \rightarrow z_{w,d}} = f_{\delta_d}\sum_{m,m' \in \text{ne}(t,t') \backslash d} \overline{\mu}_{m,t} \overline{\mu}_{m',t'},
\end{align}
where $m,m' \in \text{ne}(t,t') \backslash d$ contains all document pairs except the current document $d$ connected with the tags $\{t,t'\}$, respectively.
Eq.~\eqref{gamma2z} passes messages from the neighboring documents by the individual factor $\gamma_t$,
while~\eqref{delta2z} passes joint messages by the factor hyperedge $\delta_d$,
which connects multiple tag factors $\{\gamma_t, \gamma_{t'}\}$.
Therefore,
Eq.~\eqref{gamma2z} influences the word message $\mu(z_{w,d})$ through the pairwise relation across the individual tag $t$,
and~\eqref{delta2z} plays the similar role through the higher-order relation across multiple tags $\{t,t'\}$.
Similar to~\eqref{p2s1} and~\eqref{p2s2},
we replace the product operation by the sum operation in~\eqref{gamma2z} and~\eqref{delta2z}
for all neighboring input messages in order to avoid arithmetic underflow.

In the standard sum-product algorithm,
we calculate the marginal posterior probability $\mu_{z_{w,d}}$ by the product of all input messages according to Fig.~\ref{ttm}A.
However,
the direct product is not flexible to balance the messages from factors $\theta_d$, $\gamma_t$ and $\delta_d$ in Fig.~\ref{ttm}A.
Conceivably,
the message $\mu_{\theta_d \rightarrow z_{w,d}}$ measures the topic labeling influence within the document $d$,
the message $\mu_{\gamma_t \rightarrow z_{w,d}}$ captures the influence from neighboring documents by pairwise relations,
and the message $\mu_{\delta_d \rightarrow z_{w,d}}$ plays the similar role but through higher-order relations.
Because these three messages are in the document level,
we balance them by a simple convex combination,
and rewrite~\eqref{f2v} as
\begin{align} \label{f2v2}
&\mu(z_{w,d}) \propto \bigg[(1-\omega_1-\omega_2)\mu_{\theta_d \rightarrow z_{w,d}} \notag \\
& + \omega_1 \sum_{t \in \text{ne}(d)}\mu_{\gamma_t \rightarrow z_{w,d}} +
\omega_2\mu_{\delta_d \rightarrow z_{w,d}}\bigg] \times \mu_{\phi_w \rightarrow z_{w,d}},
\end{align}
where $\omega_1 \ge 0,\omega_2 \ge 0, \omega_1 + \omega_2 \le 1$ are the weights to balance three messages from factors $\theta_d$, $\gamma_t$ and $\delta_d$.
In~\eqref{f2v2},
we sum the messages $\mu_{\gamma_t \rightarrow z_{w,d}}$ in terms of all individual tag $t$ attached to the document $d$,
which accumulates the influence from all attached tags.
Obviously,
Eq.~\eqref{f2v2} shows that the current word message is regularized by both pairwise and higher-order relations of tagged documents.
When $\omega_1, \omega_2 = 0$,
Eq.~\eqref{f2v2} reduces to~\eqref{f2v},
so that TTM becomes LDA without tag information.
When $\omega_3 = 0$,
we depict only pairwise relations between tagged documents.
Automatic estimating the best weights in TTM requires further studies in future work.
In this paper,
we manually tune these weights based on the training data sets.

\begin{figure*}[t]
\centering
\includegraphics[width=1.0\linewidth]{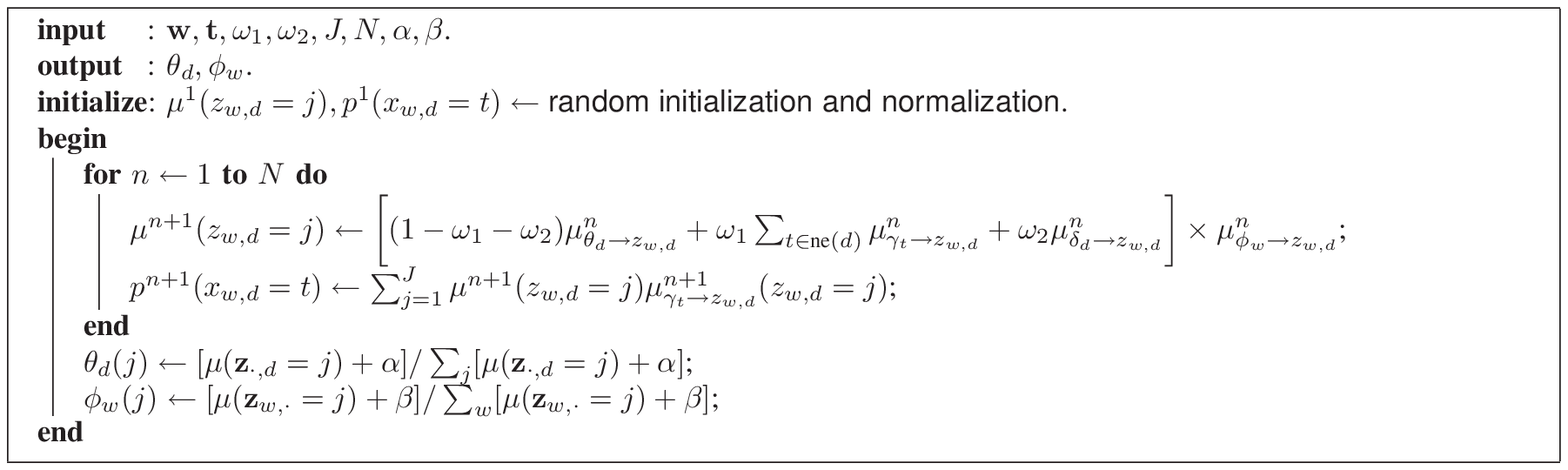}
\caption{The loopy belief propagation for learning TTM.}
\label{BPTTM}
\end{figure*}

The inference and parameter estimation equations for TTM are almost the same as those for LDA except that
the update equation of message $\mu(z_{w,d})$ is replaced by~\eqref{f2v2}.
Fig.~\ref{BPTTM} summarizes the loopy BP algorithm for learning TTM.
At each learning iteration,
we need to estimate both the pairwise and higher-order topic structural dependencies using~\eqref{pairwise} and~\eqref{3order},
so that the computational cost of learning TTM is $\mathcal{O}(JDWL)$,
where $L$ is the total number of pairwise and higher-order relations among tagged documents and images.

\section{Experimental Results} \label{s5}

\subsection{Data Sets}

\begin{table}[t]
\centering
\caption{Summarization of four data sets}
\begin{tabular}{|c|c|c|c|c|c|c|} \hline
Data sets   &$D$         &$T$      &$W$      &$I_d$          &$W_d$         &$T_d$ \\ \hline \hline
CORA        &$2410$      &$2480$   &$2961$   &$57$           &$43$          &$3.0$            \\
MED         &$2317$      &$8906$   &$8918$   &$104$          &$66$          &$5.8$            \\
C5K         &$5000$      &$371$    &$128$    &$11970$        &$90$          &$3.5$          \\
C30K        &$31695$     &$1035$   &$128$    &$11970$        &$105$         &$3.6$          \\ \hline
\end{tabular}
\label{datasets}
\end{table}

We use four data sets of tagged documents and images:
\begin{itemize}
\item
CORA~\cite{McCallum:00} and MEDLINE (MED)~\cite{Zhu:09a}:
The former contains abstracts from the Cora research paper search engine in machine learning area,
and the latter contains abstracts from the MEDLINE biomedical paper search engine.
We use the author names as the tags for each paper.
CORA documents can be classified into $7$ major categories,
and MED documents fall broadly into $4$ categories.
\item
COREL5K (C5K) and COREL30K (C30K)~\cite{Duygulu:02}:
They originate from the Corel stock photograph collection.
They contain all kinds of images,
ranging from natural scenes to people portraits or sports photographs.
Each image is associated with manually labeled tags that depict the main objects appearing in the picture.
We use the colored pattern appearance model (CPAM)~\cite{Qiu:02} to represent each image as a bag of visual words.
A sliding window decomposes the image into $11970$ visual words of $4 \times 4$ tile,
which are then mapped to one of $128$ word vocabulary indexes built by the CPAM from lots of image patches using vector quantization.
\end{itemize}
Table~\ref{datasets} summarizes the statistics of four data sets,
where $D$ is the total number of documents,
$T$ is the total number of tags,
$W$ is the vocabulary size,
$I_d$ is the average number of words per document,
$W_d$ is the average vocabulary size per document,
and $T_d$ is the average number of tags per document.

\subsection{Performance of Tag-Topic Models}

In the following experiments,
we randomly divide the entire CORA and MED documents into training ($80\%$) and test ($20\%$) sets.
For C5K,
we use the same training and test set partition of~\cite{Duygulu:02},
in which $4500$ images constitute the training set and the remaining $500$ images constitute the test set.
For C30K,
we randomly partition the entire images into training ($90\%$) and test ($10\%$) sets.
We manually tune the weights $\omega_1$ and $\omega_2$ in~\eqref{f2v2} based on the perplexity~\cite{Blei:03} for training data.
When $\omega_1 = 0.2, \omega_2 = 0$,
we refer to this TTM as TTM-P for only pairwise relation modeling.
When $\omega_1 = 0.1, \omega_2 = 0.05$,
we refer to this TTM as TTM-H for both pairwise and higher-order relation modeling.
Through the comparative study between TTM-P and TTM-H,
we can explore the effectiveness of modeling higher-order topic interactions among tagged documents and images.

We compare TTM with three current state-of-the-art topic models such as RTM (Fig.~\ref{ptm}B) with the exponential link
probability function~\cite{Chang:10}\footnote{\url{http://cran.r-project.org/web/packages/lda/}},
ATM (Fig.~\ref{ptm}C)~\cite{Rosen-Zvi:04}\footnote{\url{http://psiexp.ss.uci.edu/research/programs_data/toolbox.htm}},
and L-LDA (Fig.~\ref{ptm}D)~\cite{Ramage:09}\footnote{\url{http://nlp.stanford.edu/software/tmt/tmt-0.3/}}
using the same experimental settings.
As discussed in Section~\ref{s2},
the above benchmark topic models are able to handle pairwise relations between tagged documents.
In contrast,
TTM additionally considers higher-order relations induced by connected tags among documents.
Because L-LDA is a supervised topic model,
we only compare TTM with L-LDA in the tag recommendation task.
In all experiments,
we assume that the tags are unobserved in test data,
and use the estimated topic distributions from training data
to predict words, links, tags as well as class labels for documents in the test set.

\subsubsection{Word Prediction}

\begin{figure*}[t]
\centering
\includegraphics[width=1.0\linewidth]{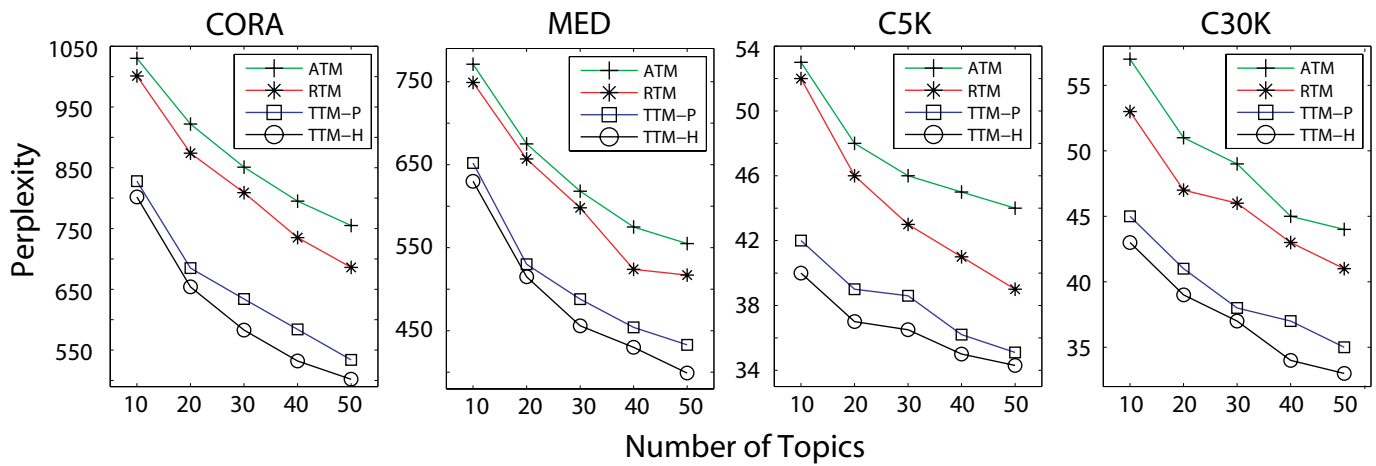}
\caption{Comparison of word prediction performance measured by the perplexity on test set.}
\label{perp}
\end{figure*}

The word prediction task is to evaluate the likelihood that the learned topic distributions generate the unseen test data.
Fig.~\ref{perp} compares the test set perplexity of ATM, RTM, TTM-P and TTM-H.
The lower perplexity corresponds to the higher likelihood that the learned topics can generate the unseen test set.
For all data sets,
TTM-H consistently achieves the lowest perplexity in different topics,
which implies the best generalization ability to predict words in unseen test sets.
Unlike RTM,
ATM does not explicitly model the pairwise topic representations between tagged documents and images,
so that it insufficiently benefit from the rich relational information for regularizing topic distributions.
On the other hand,
RTM estimates the link probability function for all document pairs connected by different tags,
while TTM-P estimates the tag-specific pairwise relations using~\eqref{pairwise}.
As a result,
TTM-P has the potential to capture the subtle topic structural dependencies between documents or images through specific tags.
Fig.~\ref{perp} shows that TTM-P achieves almost $15\%$ reduction on average in perplexity compared with ATM and RTM.
Furthermore,
TTM-H gains on average $5\%$ reduction of perplexity as compared with TTM-P,
which indicates that joint influence through higher order relations may paly positive roles in topic distribution regularization.
Although TTM-H has a higher computational complexity than TTM-P,
it is worth gaining a better word prediction performance in real-world applications.
Generally,
the predictive perplexity decreases as the number of topics increases,
which implies that the more latent topics provide the higher likelihood to predict words.

\begin{figure*}[t]
\centering
\includegraphics[width=1.0\linewidth]{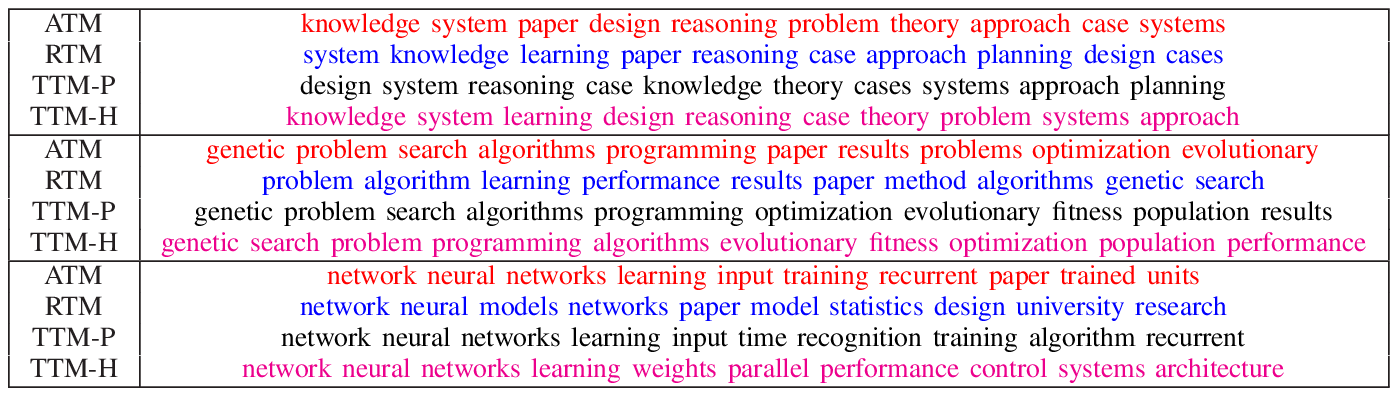}
\caption{Each row shows one topic (top ten words) of ATM, RTM, TTM-P and TTM-H on the CORA training set.
}
\label{topic}
\end{figure*}

To measure the interpretability of a topic model,
the {\em word intrusion} and {\em topic intrusion} are proposed to involve subjective judgements~\cite{Chang:09b}.
The basic idea is to ask volunteer subjects to identify the number of word intruders in the topic as well as the topic intruders in the document,
where intruders are defined as inconsistent words or topics based on prior knowledge of subjects.
Due to lack of volunteer subjects,
Fig.~\ref{topic} shows only three consistent topics with top ten words on the CORA training set for qualitative evaluation.
We see that most topics share similar words with different ranking orders.
Nevertheless,
both ATM and RTM extract first two topics that contain the word intruder ``paper",
and RTM even extracts three word intruders ``design", ``research" and ``university" in the third topic.
Obviously,
both TTM-P and TTM-H show much better interpretability at least for the top ten words,
which do not contain irrelevant common words such as ``paper".
Moreover,
TTM-H is slightly better than TTM-P in that it has a more natural word ranking order in each topic.

\subsubsection{Link Prediction}

The link prediction task is to predict if two documents share the same tag.
The natural real-world application of link prediction is to suggest tags of a document to a linking document.
If the tags are author names,
we may use the link prediction to find reviewers or collaborators for the linking document.
Also,
the link prediction can help retrieve related documents with similar tags.
The effectiveness of these applications depend highly on the link prediction accuracy.
We define the link prediction as a binary classification problem.
We use the Hadmard product of a pair of document topic proportions as the link feature,
and train an SVM~\cite{Chang:11} to decide if there is a link between them.
We evaluate link prediction performance using the same number of linking/non-linking training and test samples.

\begin{figure*}[t]
\centering
\includegraphics[width=1.0\linewidth]{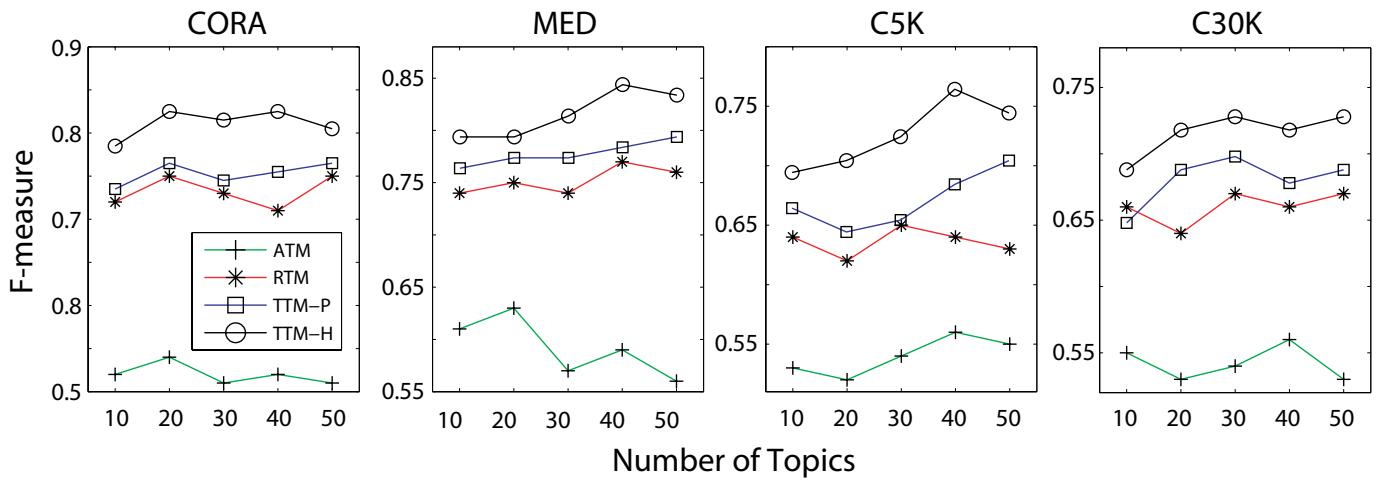}
\caption{Comparison of link prediction performance based on the Hadmard product of document topic proportions.
}
\label{link}
\end{figure*}

Fig.~\ref{link} compares the F-measure of link prediction.
Because ATM does not encode the Hadmard link features of pairs of documents,
its prediction results are almost random guess with F-measure close to $0.5$ for all data sets.
In contrast,
RTM shows a significantly better link prediction performance using the generalized linear models estimated from the link features,
which efficiently differentiate links from non-links.
For all data sets,
TTM-P deviates slightly from RTM because both TTM-P and RTM encode only pairwise relations of tagged documents.
However,
TTM-H outperforms RTM around $8\%$ F-measure for link prediction.
One possible reason is that TTM-H incorporates much richer higher-order topic structural dependencies so that
it makes the topic proportions of documents sharing tags more differentiable from those documents without sharing tags.
Interestingly,
F-measure does not always increase as the number of topics increases.
Although more latent topics can predict unseen words better as shown in Fig.~\ref{perp},
they cannot consistently enhance the link prediction performance on the test set as shown in Fig.~\ref{link}.
This phenomenon suggests that the content similarity between documents alone cannot completely account for the link information.
Additional information such as partially observed links of some documents may help for a better link prediction performance.

\subsubsection{Document Classification}

Document classification partitions a set of documents into several mutually exclusive categories.
Topic models can be used as a dimensionality reduction method to reduce the high-dimensional word vector space for classification~\cite{Blei:03}.
We may use the document topic proportions as the reduced feature vectors and study their discriminative ability in document classification.
To this end,
we train SVMs on the document topic proportions given class labels,
and compare the document classification accuracy on the test set.
In CORA,
we randomly select $100$ training samples for each of the seven categories.
In MED,
we randomly select $200$ training samples for each of the five categories.
In C5K and C30K,
we choose four tags as class labels:
sky, water, trees, and people.
We use those images associated with only one of four tags for training purposes.
In C5K,
we randomly select $300$ training samples for each class.
In C30K,
we randomly select $1500$ training samples for each class.
The remaining documents and images are test samples.

\begin{figure*}[t]
\centering
\includegraphics[width=1.0\linewidth]{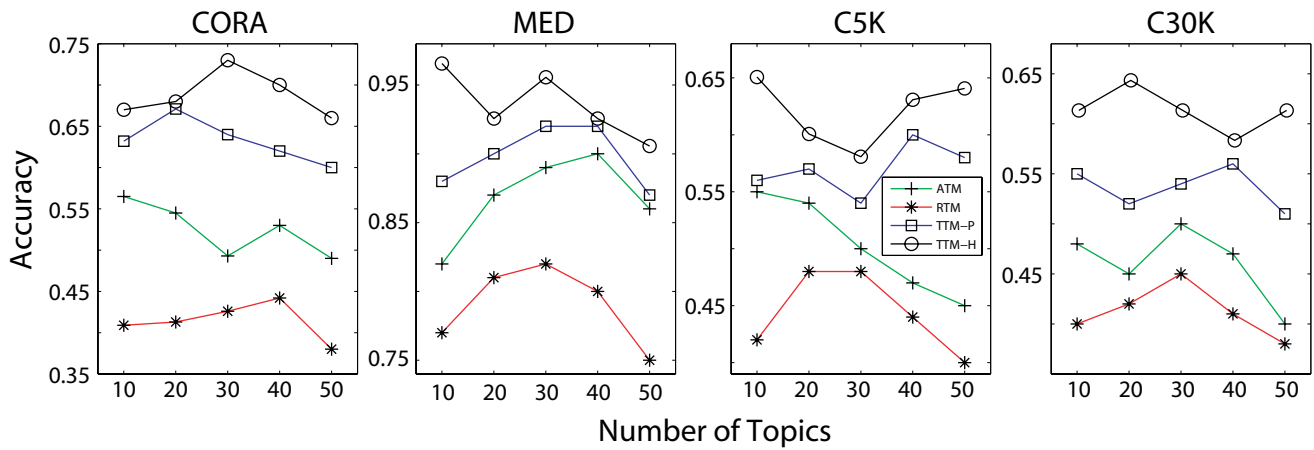}
\caption{Comparison of document classification accuracy based on document topic proportions.
}
\label{doc}
\end{figure*}

Fig.~\ref{doc} shows the classification accuracy based on low-dimensional document topic proportions.
We see that ATM generally outperforms RTM,
which is inconsistent with their word prediction performance in Fig.~\ref{perp}.
The reason may lie in that RTM treats sharing tags as equal links,
but in reality different tags may encode different topic structural dependencies between documents.
Thus,
RTM may erroneously encourage the topic smoothness of documents through different tags,
which often has the close correspondence with the class labels of documents,
especially when tags are used as class labels for C5K and C30K.
In contrast,
TTM-P relaxes the limitation in RTM by encouraging the smoothness of document topic proportions using the tag-specific pairwise relation modeling.
Furthermore,
TTM-H still outperforms TTM-P with $6\%$ higher classification accuracy on average
by forcing tag-specific smoothness constraint through both pairwise and higher-order relations.
Image classification performance on C5K and C30K is generally worse than that on CORA and MED,
partly because the tags tend to describe individual image components,
which are not exactly equivalent to class labels that describe the global image contents.
Similar to the link prediction task,
the more latent topics does not enhance the overall document classification performance.

\subsubsection{Tag Recommendation}

Tag recommendation is a multi-label classification problem that suggests a set of tags to query documents or images,
which has found many real-world applications such as credit attribution~\cite{Ramage:09},
expert finding~\cite{Mimno:07} and image annotation~\cite{Duygulu:02,Carneiro:07}.
Due to lack of benchmark data to evaluate the expert finding performance,
we focus on tag recommendation for image annotation in this section.
We propose an TTM-based tag recommendation system including two SVMs:
\begin{enumerate}
\item
Each tag is a class label.
We train a multiclass SVM called $svm_1$ to classify the image topic proportions into $T$ tags,
where the training samples are images associated with each tag.
Some images may have more than one tag and will be used as training samples for multiple tags.
For each training sample $d$,
$svm_1$ predicts a vector of likelihoods $\mathbf{p}^d = [p_1,\dots,p_t,\dots,p_T]$ for all tags.
\item
We also train a total of $T$ binary SVMs called $svm_2$ for all tags.
For the tag $t$,
the positive sample is the tagged image $d$ with the feature vector $\mathbf{p}_t^d = [p_t,p_{t' \in \text{ne}(t)}]$ predicted by $svm_1$,
where $\text{ne}(t)$ is a set of connected tags of the tag $t$.
This feature encodes information of connected tags for robust prediction.
To balance the training data for each tag,
we choose the same number of positive/negative samples.
For each training sample $d$,
$svm_2$ predicts a vector of likelihoods $\boldsymbol{\mu}_t^d = [\mu_{t,i}^d], \sum_i\mu_{t,i}^d=1, i = \{1,2\}$,
where $\mu_{t,1}^d$ is the likelihood that the tag $t$ is recommended to $d$.
\item
For the test image $d$,
we use $svm_1$ to predict its likelihoods $\mathbf{p}^d$ to all $T$ tags.
Then,
we use $svm_2$ to predict $\boldsymbol{\mu}_t^d, 1 \le t \le T$ for all tags.
To balance the prediction results of $svm_1$ and $svm_2$,
we linearly combine the two likelihoods $y = \omega p_t^d + (1-\omega)\mu_{t,1}^d$
by the best mixture weight $\omega = 0.25$ estimated from the training set.
We follow the standard image annotation evaluation protocol~\cite{Duygulu:02,Carneiro:07},
and suggest top five tags to the query image with highest $y$.
\end{enumerate}
In this system,
$svm_1$ uses only image content information to suggest tags,
while $svm_2$ uses connected tags to refine the tag recommendation result.
The basic idea is that if the tag $t$ is suggested to the image $d$,
its connected tags also have a high likelihood to be suggested.

The performance measures for image tag recommendation include recall and precision rates per tag~\cite{Duygulu:02,Carneiro:07}.
More specifically,
for a given tag,
let $N_h$ be the number of images in the test set that are labeled with this tag by human,
$N_s$ be the number of images in the test set that are labeled with tag by the tag recommendation system,
and $N_c$ be the number of images that the system gives correct tag recommendation.
The recall and precision rates are defined as
$recall = N_c/N_h$ and $precision = N_c/N_s$.
We also evaluate the coverage rate $Rate^+$ of recommended tags,
which is calculated as the number of tags with positive recall divided by the total number of tags in the test set.
The higher $Rate^+$ implies a better generalization ability that
can achieve relative high recall and precision rates on a large set of tags.

\begin{table}[t]
\centering
\caption{Comparison of Tag Recommendation}
\begin{tabular}{|c|c|c|c|}
\hline
C5K         &Recall      &Precision      &$Rate^+$  \\ \hline \hline
TTM-H       &$0.33$      &$0.22$         &$53.85\%$    \\
TTM-P       &$0.30$      &$0.21$         &$49.29\%$    \\
L-LDA       &$0.26$      &$0.14$         &$50.77\%$     \\
SML         &$0.29$      &$0.23$         &$52.69\%$ \\ \hline \hline
C30K        &Recall      &Precision      &$Rate^+$  \\ \hline \hline
TTM-H       &$0.22$      &$0.13$         &$45.89\%$    \\
TTM-P       &$0.20$      &$0.11$         &$42.00\%$    \\
L-LDA       &$0.11$      &$0.07$         &$30.40\%$     \\
SML         &$0.21$      &$0.12$         &$44.63\%$ \\ \hline
\end{tabular}
\label{corel}
\end{table}

Table.~\ref{corel} compares TTM-H and TTM-P with two state-of-the-art tag recommendation methods L-LDA~\cite{Ramage:09} and SML~\cite{Carneiro:07}.
With the similar coverage rate,
TTM-H provides the competitive image annotation performance with SML.
Although L-LDA shows the comparable or better tag recommendation performance than SVM for tagged web pages,
it does not show clear advantages in image annotation problem especially on the C30K data set.
Indeed,
L-LDA does not use the connected tag information from training data,
which play major roles to rule out many false positives to enhance the average precision.
We see that TTM-H still outperforms TTM-P,
which is consistent with its superior document classification performance as shown in Fig.~\ref{doc}.
Furthermore,
the more latent topics does not improve the tag recommendation performance,
so that we show only the best results of TTM-H and TTM-P when the number of latent topics $J = 20$.

\section{Conclusions} \label{s6}

This paper has presented TTM and discussed its effectiveness in encoding smoothness pairwise and higher-order topic interactions among tagged documents and images.
Within the MRF framework,
TTM allows the efficient loopy BP algorithm for inference and parameter estimation.
On four large-scale data sets,
TTM consistently outperforms current state-of-the-art topic models,
such as ATM, RTM and L-LDA,
in several real-world text and image mining applications.

Furthermore,
we observe that the higher-order relations also exist in many important computer vision and text mining applications.
For example,
the unsupervised activity perception in crowded and complicated scenes involves lots of higher-order interactions of multiple agents,
which can be encoded in topic models for discovering more specific motion patterns.
Another example is tracking historical topics from time-stamped documents.
We speculate that the higher-order temporal topic interactions may characterize some specific long-range topic evolution patterns,
which can be also studied in our future work.

\section*{Acknowledgements} \label{s7}

This work is supported by NSFC (Grant No. 61003154)
and the Shanghai Key Laboratory of Intelligent Information Processing, China (Grant No. IIPL-2010-009).


\end{document}